\documentclass{article}
\usepackage{ijcai23}

\usepackage{color}
\definecolor{green}{rgb}{0.0,0.75,0.0}
\definecolor{red}{rgb}{0.95,0.1,0.1}
\definecolor{blue}{rgb}{0.1,0.1,0.95}
\definecolor{cyan}{rgb}{0.0,0.85,0.85}

\usepackage{times}
\usepackage{soul}
\usepackage{url}
\usepackage[hidelinks]{hyperref}
\usepackage[utf8]{inputenc}
\usepackage[small]{caption}
\usepackage{graphicx}
\usepackage{amsmath}
\usepackage{amsthm}
\usepackage{booktabs}
\usepackage{algorithm}
\usepackage{algorithmic}
\usepackage[switch]{lineno}
\usepackage{amssymb}
\usepackage{multirow}
\usepackage{bbding}

\usepackage[capitalize]{cleveref}
\crefname{section}{Sec.}{Secs.}
\Crefname{section}{Section}{Sections}
\Crefname{table}{Table}{Tables}
\crefname{table}{Tab.}{Tabs.}
\Crefname{figure}{Figure}{Figures}
\crefname{figure}{Fig.}{Figs.}
\crefname{equation}{Equ.}{Equs.}

\title{Physics-Guided Human Motion Capture with Pose Probability Modeling}

\author{
Jingyi Ju$^{1,2}$\footnotemark[1]
\and
Buzhen Huang$^{1,2}$\footnotemark[1]\and
Chen Zhu$^{1,2}$\and
Zhihao Li$^{3}$\And
Yangang Wang$^{1,2}$\footnotemark[2]
\affiliations
$^1$Southeast University\\
$^2$Key Laboratory of Measurement and Control of Complex Systems of Engineering, Ministry of Education, Nanjing, China\\
$^3$Huawei Noah’s Ark Lab\\
\emails
\{jingyiju, hbz, yangangwang\}@seu.edu.cn,
zc1213856@163.com,
zhihao.li@huawei.com
}

\begin{document}

\maketitle

\begin{abstract}
    Incorporating physics in human motion capture to avoid artifacts like floating, foot sliding, and ground penetration is a promising direction. Existing solutions always adopt kinematic results as reference motions, and the physics is treated as a post-processing module. However, due to the depth ambiguity, monocular motion capture inevitably suffers from noises, and the noisy reference often leads to failure for physics-based tracking. To address the obstacles, our key-idea is to employ physics as denoising guidance in the reverse diffusion process to reconstruct physically plausible human motion from a modeled pose probability distribution. Specifically, we first train a latent gaussian model that encodes the uncertainty of 2D-to-3D lifting to facilitate reverse diffusion. Then, a physics module is constructed to track the motion sampled from the distribution. The discrepancies between the tracked motion and image observation are used to provide explicit guidance for the reverse diffusion model to refine the motion. With several iterations, the physics-based tracking and kinematic denoising promote each other to generate a physically plausible human motion. Experimental results show that our method outperforms previous physics-based methods in both joint accuracy and success rate. More information can be found at \url{https://github.com/Me-Ditto/Physics-Guided-Mocap}.

\end{abstract}

\section{Introduction}\label{sec:Introduction}

\begin{figure}
    \begin{center}
    \includegraphics[width=1.0\linewidth]{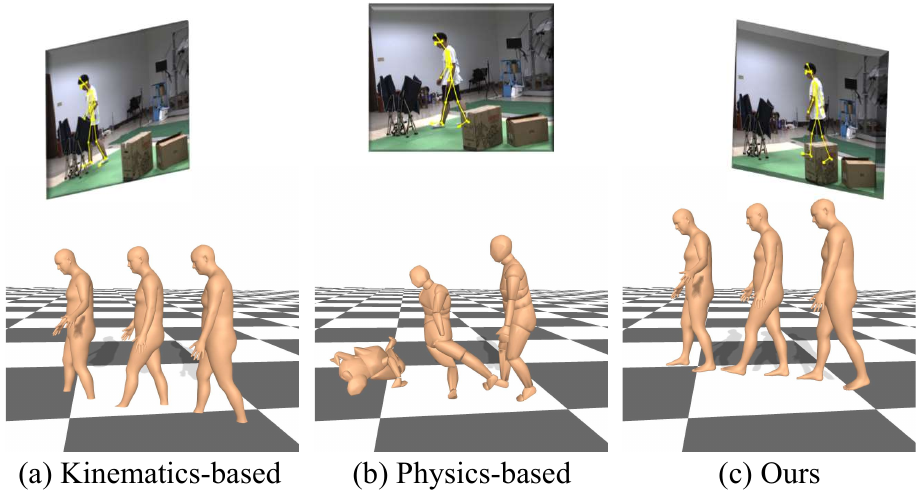}
    \end{center}
    \caption{The kinematics-based approaches~(a) suffer from artifacts, and dynamics-based works~(b) encounter tracking failure, while our method~(c) can reconstruct physically plausible human motion with a high success rate.}
\label{fig:teaser}
\end{figure}

Human motion capture is a fundamental task in sports broadcasting, human behavior understanding, and virtual reality, which require the accurate perception of human pose, position, and contact. Previous kinematics-based works~\cite{kocabas2020vibe,arnab2019exploiting,kanazawa2019learning,rempe2021humor} neglect the physical laws when exploring human motion capture from monocular videos and images. As shown in~\cref{fig:teaser}~(a), even state-of-the-art monocular kinematics-based motion capture suffers from artifacts (\eg, floating, foot sliding, and ground penetration) due to the occlusion and depth ambiguity.

To tackle this problem, recent works introduce physical laws in human motion capture. The optimization-based framework~\cite{huang2022neural,shimada2020physcap} relieves artifacts by solving a highly-complex formulation. Others rely on Reinforcement Learning~(RL)~\cite{yuan2021simpoe,luo2021dynamics} with non-differentiable physics simulators to obtain a physically plausible human motion. However, these methods all use a physical character to track the kinematic motion, and the physics is treated as a post-processing module. The noises in the kinematic motion always lead to tracking failure and thus result in a low success rate.

To address these limitations, \textbf{our key idea is to employ physics as denoising guidance in a reverse diffusion process to reconstruct physically plausible human motion from modeled pose probability distributions.} Thus, physics can guide the denoising process to progressively improve the motion quality. Nonetheless, its implementation still faces several technical obstacles. First, since highly-precise simulators are non-differentiable, the physics module cannot be incorporated into the network to provide explicit gradients to optimize human motions. In the generative task, PhysDiff~\cite{yuan2022physdiff} directly uses a tracked motion as the input of the reverse diffusion step to avoid artifacts. However, this method also requires relatively high-quality reference motion. In addition, the strategy cannot be directly applied in motion capture since it does not consider 2D observations. Second, the denoising models~\cite{ho2020denoising,song2020denoising} always start from the standard gaussian distribution to generate a sample~\cite{li2022diffusion,ramesh2022hierarchical,zhang2022motiondiffuse,yuan2022physdiff}, which ignores the prior knowledge from image observations and may require thousands of denoising steps to reconstruct a satisfactory motion.

To fully utilize motion prior knowledge to improve the denoising efficiency, we first train a latent gaussian model based on Variational Autoencoder (VAE)~\cite{kingma2013auto}. With the trained VAE encoder, the image features are mapped to a series of gaussian distributions to reflect the 3D probabilistic motion. The modeled probabilistic distributions can be used as good initial values to facilitate the denoising process. To alleviate the artifacts in the reconstructed motion, we further propose a physics module in the reverse diffusion model to provide implicit guidance for the denoising. Different from PhysDiff~\cite{yuan2022physdiff}, we utilize the discrepancies between the tracked motion and image observation to guide the reverse diffusion process in the next timestep. Specifically, we project the tracked joint positions to the 2D image plane and calculate the projection loss gradients. Then, we use the combination of the gradients and image features as a condition, and feed the tracked motion to the next reverse diffusion step. After several iterations, physics-based tracking and kinematic denoising can promote each other to obtain a physically plausible human motion. The main contributions of this paper are as follows:

\begin{itemize}
    \item We construct a physical guidance to combine the physics and image observations for the reverse diffusion process to progressively promote a physically plausible human motion capture.
    \item We propose a VAE-based latent gaussian distribution to facilitate the reverse diffusion process with motion prior knowledge.
    \item We incorporate physics and kinematics in the same framework to improve joint accuracy and success rate for physics-based human motion capture.

\end{itemize}

\begin{figure*}
    \begin{center}
    \includegraphics[width=1.0\linewidth]{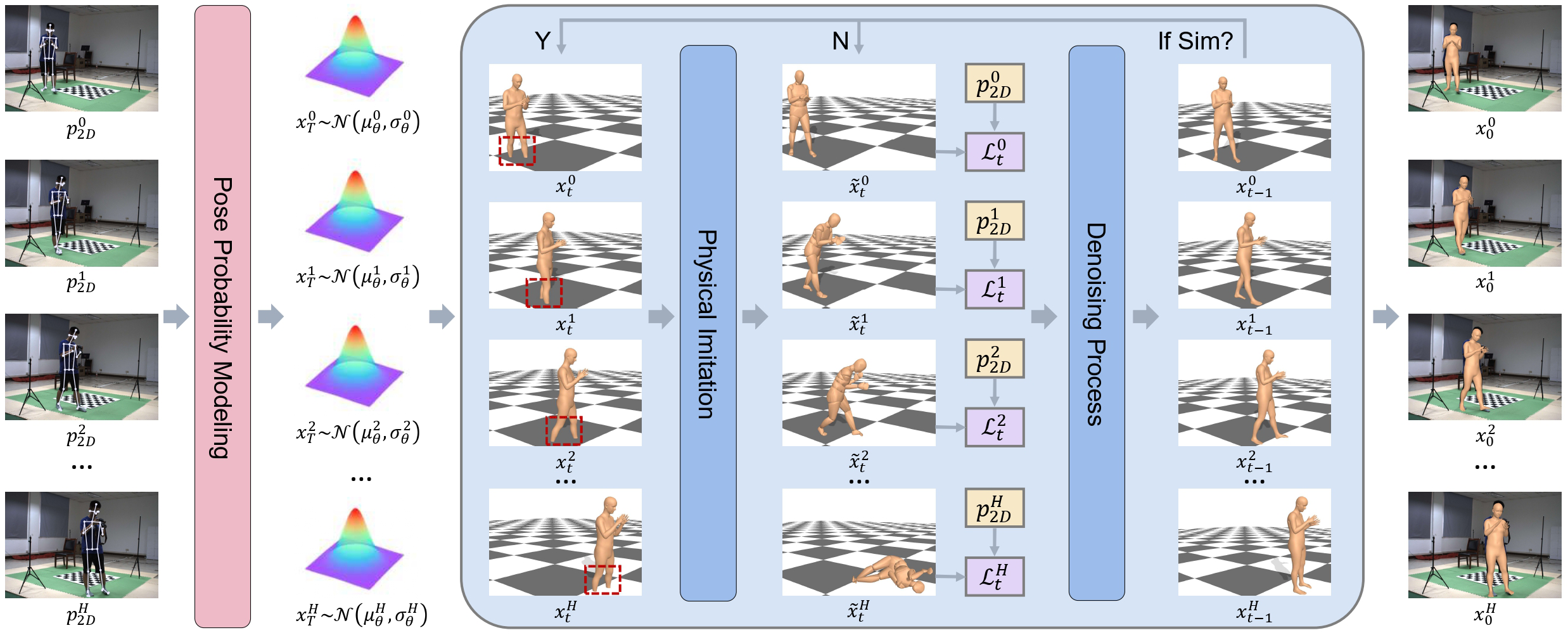}
    \end{center}
    \caption{We formulate the physics-based motion capture as a reverse diffusion process. Given images and 2D poses estimated from off-the-shelf 2D pose detector, our method first regresses a series of gaussian distributions $\mathcal{N} \left(\mu_{\theta},\sigma_{\theta} \right)$ from color images with a trained VAE encoder. We then sample a human motion from the encoded distributions and use it as the initial value for the diffusion model. To improve the physical plausibility and tracking success rate, we further propose a physical guidance that combines physics and 2D observations to guide the denoising. After several denoising steps, the physically plausible human motion can be obtained.}
\label{fig:pipeline}
\end{figure*}

\section{Related Work}\label{sec:relatedwork}

\subsection{Kinematics-Based Motion Capture}
Previous monocular kinematics-based motion capture leverages 3D pose estimation~\cite{li2021hybrik,li2022cliff,song2020human} for each frame to construct the human motion. They cannot obtain the temporal information among frames, which leads to obvious jittering. Recent works~\cite{kocabas2020vibe,luo20203d} exploit the temporal context of human motion for better temporal consistency. These methods encounter global inconsistency since they can only produce a root-relative motion. Several approaches~\cite{arnab2019exploiting,xiang2019monocular} adopt smooth priors over time to reduce jittering. However, the smooth constraints may result in over-smooth and footskate. Since the aforementioned methods do not consider depth ambiguity and occlusion from monocular motion capture, recent diffusion-based works~\cite{gong2022diffpose,choi2022diffupose,holmquist2022diffpose} model the uncertainty of 2D-3D lifting for 3D pose estimation. Although the aforementioned approaches achieve great performance on kinematic metrics, they still encounter artifacts since they ignore the physics laws.

\subsection{Physics-Based Motion Capture}
To relieve the artifacts of human motion capture,~\cite{wei2010videomocap,vondrak2012video,zell2017joint,shimada2020physcap,shimada2021neural,li2022estimating,yi2022physical} adopt optimization to obtain the physical forces to induce the human motion, which results in high approximation errors since they do not utilize the highly-precise non-differentiable simulators. While Neural Mocon~\cite{huang2022neural} generates a learned motion distribution for sampling-based motion control with supervision from a non-differentiable simulator. Others~\cite{yuan2021simpoe,luo2021dynamics,yuan2020residual,yu2021human,peng2018sfv} use Reinforcement Learning (RL) with non-differentiable simulators to obtain physically plausible human motion. However, all the approaches adopt a strong assumption that the reference motion is accurate. Luo~\etal~\cite{luo2022embodied} adopts current character state and environmental cues to promote tracking when the reference pose is unreliable. The accuracy of the estimated motion relies on the 2D observations. Since all these approaches implement physics as a post-processing process, the character always fails to track due to the noises in the reference motion. In contrast, we propose a physics module in the reverse diffusion model to provide implicit guidance for the denoising. In this case, the physics-based tracking and kinematic denoising can promote each other for physically plausible human motion capture.

\subsection{Multi-Hypothesis Estimation}

Human motion capture from monocular videos and images is an ill-posed problem, for which directly regressing a determinate solution may be inaccurate~\cite{kocabas2020vibe,arnab2019exploiting,kanazawa2019learning}. Multi-hypothesis methods~\cite{huang2022object,li2019generating,jahangiri2017generating} are proposed to represent the uncertainty of 2D-3D lifting. Recently, Sharma~\etal~\cite{sharma2019monocular} adopted a conditional VAE to predict 3D pose candidates conditioned on detected 2D poses. Wehrbein~\etal~\cite{wehrbein2021probabilistic} employs normalizing flow to model the posterior distribution of 3D poses. However, it is difficult to select the best 3D pose from multi-hypothesis. Unlike these works, we propose a physics module to provide implicit guidance for the denoising process to promote a more accurate human motion.

\section{Method}\label{sec:method}

We aim to reconstruct the physically plausible 3D human motion from monocular videos. We first design a latent gaussian distribution encoded from image features to provide initial values to facilitate the reverse diffusion process~(\cref{sec:distribution}). To alleviate the artifacts in the reconstructed motion, we further propose a physics module to combine the physics and observed 2D poses to guide the denoising of reverse diffusion~(\cref{sec:physical diffusion}). With the constructed framework, the physics and kinematics can progressively promote each other to obtain physically plausible motions with a high success rate~(\cref{sec:motion capture}).

\subsection{Preliminaries}\label{sec:preliminaries}
\paragraph{Motion Representation.}
The 2D poses with corresponding confidence detected by AlphaPose~\cite{fang2017rmpe} from color images $I^{1:H}$ are defined $\mathcal{P}_{2D}=\left\{p_{2D}^h \in \mathbb{R}^{J \times 3} \right\}_{h=1}^H$, where $H$ is the length of motion and $J$ is the number of joints. We adopt SMPL model~\cite{loper2015smpl} to represent the kinematic human motion $x^{1:H} = \left\{x^h \right\}_{h=1}^{H}$, where $x$ denotes the parameters of human pose $\theta$ in 6D representation~\cite{zhou2019continuity}. We also regress SMPL shape $\beta$ and translation $\tau$. The physics-based human motion is defined as $\tilde{x}^{1:H}$.

\paragraph{Physics-Based Tracking.}
We briefly introduce the physical imitation. The character in the physics engine is created based on the SMPL kinematic tree and estimated body shape $\beta$. We construct convex hulls to approximate mesh~\cite{luo2021dynamics}, which can be simulated in MuJoCo~\cite{todorov2012mujoco} and shares the same pose parameters with SMPL model. The physical imitation aims to control the character to track the reference motion $x^{1:H}$. With a trained policy $\pi(a^h \mid s^h, x^{h+1})$, we can sample an action $a^h$ according to the current state $s^h$ and reference pose $x^{h+1}$ to control the character to move to the next state $s^{h+1}$. The state $s^h = (\tilde{x}^h, \dot{\tilde{x}}^h)$ contains the character’s current pose $\tilde{x}^h$ and joint velocity $\dot{\tilde{x}}^h$. Finally, the physically plausible human motion $\tilde{x}^{1:H}$ can be obtained from the simulated character state. More details can be found in the supplementary material.

\paragraph{Diffusion Model.}
The diffusion model~\cite{tevet2022human} can generate target data from a simple noise distribution under a condition. Specifically, the forward diffusion process gradually adds infinitesimal gaussian noise $\epsilon$ on the data $x_t^{1:H} \sim q(x_t^{1:H})$ at timestep $t$, which is formulated as:
\begin{equation}\label{equation:forward_diffusion}
    q(x_t^{1:H} \mid \bar{x}_0^{1:H}) = \sqrt{\hat{\alpha}_t}\bar{x}_0^{1:H} + \sqrt{1 - \hat{\alpha}_t} \epsilon, \epsilon \sim \mathcal{N} \left(0, \rm{I} \right),
\end{equation}
where $\alpha_t$ is a manually designed constant hyper-parameter, and $\hat{\alpha}_t = \prod_{i=0}^t \alpha_i$, and $\bar{x}_0^{1:H}$ is the ground truth original data.
The reverse diffusion process samples an initial input from the standard gaussian distribution and progressively denoises it to the target data under the guidance of condition $c$, which is defined as:
\begin{equation}\label{equation:reverse_diffusion}
    q(x_{t-1}^{1:H} \mid x_{t}^{1:H},c) = \mathcal{N} (x_{t-1}^{1:H}; \mu_{\alpha}(x_{t}^{1:H},c),\tilde{\beta_t} \rm{I}),
\end{equation}
where $\mu_{\alpha}(x_{t}^{1:H},c)$ is the estimated mean by a neural network and $\tilde{\beta_t}$ is the variance which is calculated by the hyper-parameters $\beta_t$, $\hat{\alpha}_t$ and $\hat{\alpha}_{t-1}$.

\subsection{Probabilistic Modeling of 3D Human Motion}\label{sec:distribution}

In this work, we formulate the motion capture as a reverse diffusion process conditioned on image observations. The previous diffusion models~\cite{choi2022diffupose,gong2022diffpose} always start from the standard gaussian distribution, which requires thousands of reverse diffusion timesteps to denoise the input to produce a desired result. Although a recent work~\cite{gong2022diffpose} separately models the initial distribution for each pose coordinate, it neglects the correlation among different joints. To improve efficiency, we leverage prior knowledge from image features to provide initial values for the reverse diffusion process.

Inspired by the previous works~\cite{huang2022pose2uv,huang2021dynamic}, we first extract image features $\mathcal{F}^{1:H}$ with a backbone network, and then fed the features into a VAE model to output 3D motion $x^{1:H}$, translation $\tau$ and body shape $\beta$. Since we adopt a Gate Recurrent Unit (GRU) for the VAE encoder,  the correlated gaussian distributions $\left\{\mathcal{N} \left( \mu_\theta \left( \mathcal{F}^h \right),\sigma_\theta \left( \mathcal{F}^h \right) \right) \right\}_{h=1}^H$ in the latent space can describe the motion prior knowledge. During the training phase, we maximize the Evidence Lower Bound (ELBO) to train the model:
\begin{equation}
    \begin{array}{l}
    \log p_{\theta}\left(x^{1:H}\right) \geq \mathbb{E}_{q_{\phi}}\left[\log p_{\theta}\left(x^{1:H}\mid\mathbf{z}\right)\right] \\
    -D_{\mathrm{KL}}\left(q_{\phi}\left(\mathbf{z} \mid \mathcal{F}^{1:H}\right) \| p_{\theta}\left(\mathbf{z}\right)\right).
    \end{array}
\end{equation}
The overall loss function is:
\begin{equation}
    \mathcal{L}_{\mathrm{VAE}} = \mathcal{L}_{\mathrm{motion}} + \mathcal{L}_{\mathrm{shape}} +\mathcal{L}_{\mathrm{joint}}+ \mathcal{L}_{\mathrm{reproj}} + \mathcal{L}_{\mathrm{kl}},
\end{equation}
where $\mathcal{L}_{\mathrm{motion}}$ and $\mathcal{L}_{\mathrm{shape}}$ are used to supervise the estimated human motion and shape:
\begin{equation}
    \mathcal{L}_{\mathrm{motion}}=\sum_{h=1}^{H}\left\|x^{h}-{\bar{x}}^{h}\right\|^{2},
\end{equation}

\begin{equation}
    \mathcal{L}_{\mathrm{shape}}=\left\|\beta-\mathbf{\bar{\beta}}\right\|^{2},
\end{equation}
where ${\bar{x}}^{h}$ and $\mathbf{\bar{\beta}}$ are ground truth motion and shape. 
\begin{equation}\label{equation:joint}
    \mathcal{L}_{\mathrm{joint}}=\sum_{h=1}^{H}\left\|J_{3D}^h-{\bar{J}_{3D}}^{h}\right\|^{2},
\end{equation}
where $\bar{J}_{3D}^h$ and $J_{3D}^h$ are the ground truth and predicted 3D joint positions generated from the corresponding motion.
We also follow CLIFF~\cite{li2022cliff} to supervise the predicted joints in the original camera coordinates with detected 2D poses:
\begin{equation}\label{equation:reproj}
    \mathcal{L}_{\mathrm{reproj}}=\frac{1}{H} \sum_{h=1}^{H}\left\| \Pi\left(J_{3D}^h\right) - p_{2D}^h \right\|_2^2,
\end{equation}
where $\Pi$ denotes the projection operation. 
\begin{equation}
    \mathcal{L}_{kl}=K L(q_\phi(\mathbf{z}\mid \mathcal{F}^{1:H}) \| \mathcal{N}(0, \rm{I})),
\end{equation}
which is used to push the output of the encoder to approximate the gaussian distribution.

After the training, we freeze the network parameters of the VAE encoder and use it to generate specific gaussian distributions from image features for the reverse diffusion process.

\subsection{Physics-Guided Motion Diffusion}\label{sec:physical diffusion}

Although the human motion sampled from the encoded distributions can match the image observations, it is still physically implausible due to the occlusion and depth ambiguity. Thus, we incorporate physics to refine the kinematic motion. Previous works~\cite{yuan2021simpoe,shimada2020physcap} directly use the physics module as a post-processing to track the kinematic motion. However, the noises in kinematic motions always result in tracking failure. To address the obstacle, we use physics to guide the reverse diffusion process to denoise the motion.

Unlike previous diffusion-based pose estimation~\cite{choi2022diffupose,holmquist2022diffpose}, we start from the modeled probabilistic distributions, which can provide prior knowledge to improve the efficiency of the denoising process. To train the diffusion model, we first use the trained VAE encoder to produce gaussian distributions $\mathcal{N} \left( \mu_\theta^{1:H},\sigma_\theta^{1:H} \right)$ for the diffusion framework. The difference from the standard diffusion process~\cref{equation:forward_diffusion} is that we sample the noise $\epsilon$ from the encoded distributions rather than the standard gaussian distribution. We then gradually add the sampled noises on the 3D motion $\bar{x}_0^{1:H}$ towards the uncertainty distribution $\mathcal{N} \left( \mu_\theta^{1:H} ,\sigma_\theta^{1:H} \right)$.

\begin{equation}\label{equation:my_forward_diff}
\begin{aligned}
    q(x_t^{1:H} \mid \bar{x}_0^{1:H}) & = \sqrt{\hat{\alpha}_t}\bar{x}_0^{1:H} + \sqrt{1 - \hat{\alpha}_t} \epsilon,\\ \epsilon & \sim \mathcal{N} \left( \mu_\theta^{1:H} ,\sigma_\theta^{1:H} \right).
\end{aligned}
\end{equation}

In the reverse diffusion process, we train a network to denoise the noisy motion $x_T^{1:H}$ to the original data $x_0^{1:H}$. Since the distributions have encoded prior knowledge, the noisy motion only contains a few artifacts and is still close to the real motion. Although directly applying the physics-based tracking on the noisy reference motion may fail, the discrepancies between the tracked motion and 2D poses can provide implicit guidance for the next reverse diffusion step to optimize the kinematic motion. Thus, we combine the projection loss gradient and image features as a condition, and feed the tracked motion to the network to predict the distributions of the next step.
\begin{equation}\label{equation:predict_next_data}
    q \left( x_{t-1}^{1:H} \mid \tilde{x}_{t}^{1:H}, c_t\right) = \mathcal{N} \left( x_{t-1}^{1:H}; \mu_\alpha \left( \tilde{x}_t^{1:H}, c_t \right), \tilde{\beta}_t \rm{I} \right).
\end{equation}
We follow Ramesh~\etal~\cite{ramesh2022hierarchical} to make the diffusion model to predict the target data $\check{x}_0^{1:H}$, and then construct the mean of the distribution $\mu_\alpha \left( \tilde{x}_t^{1:H}, c_t \right)$ in timestep $t-1$ according to $\check{x}_0^{1:H}$. The condition $c_t$ is a concatenated vector of image features and projection loss gradient information. The image features vector is encoded from extracted image features with a linear layer, and the gradient vector records the gradient of each frame.
\begin{equation}
    \mathcal{L}_t^h = \frac{\partial \left\| \Pi\left(\tilde{J}_{3 D}^h\right) - p_{2D}^h \right\|_2^2}{\partial\tilde{J}_{3 D}^h}.
\end{equation}
We mask the gradient for the frames that are not successfully tracked. We also fill the gradient factor with 0 when the physical guidance is not executed.

\paragraph{Training Procedure.} For the physics-based tracking, we first follow~\cite{luo2021dynamics} to train a policy $\pi$ on motion capture datasets by maximizing the expected return, and the parameters of the trained policy are fixed in the diffusion model training. To train the diffusion model, the following loss function is adopted:
\begin{equation}
    \mathcal{L} = \mathcal{L}_{\mathrm{diff}} + \mathcal{L}_{\mathrm{joint}} + \mathcal{L}_{\mathrm{reproj}}.
\end{equation}

\begin{equation}
    \mathcal{L}_{\mathrm{diff}} = \mathbb{E}_{\bar{x}_0^{1:H} \sim q\left( \bar{x}_0^{1:H}  \right),t \sim \left[ 1,T \right]} \left[ \big\| \bar{x}_0^{1:H} - F\left( \tilde{x}_t^{1:H},t,c_t \right) \big\|_2^2 \right],
\end{equation}
where $F\left( \tilde{x}_t^{1:H},t,c_t \right)$ is the output of the neural network. The functions $\mathcal{L}_{\mathrm{joint}}$ and $\mathcal{L}_{\mathrm{reproj}}$ are the same as~\cref{equation:joint} and~\cref{equation:reproj}.

\begin{table*}
    \begin{center}
        \resizebox{1.0\linewidth}{!}{
            \begin{tabular}{l|c c c c c |  c c c |  c c }
            \noalign{\hrule height 1.5pt}
            \begin{tabular}[l]{l}\multirow{2}{*}{Method}\end{tabular}
                &\multicolumn{5}{c|}{Human3.6M}  &\multicolumn{3}{c|}{3DOH}  &\multicolumn{2}{c}{3DHP} \\
                &MPJPE$\downarrow$ &PA-MPJPE$\downarrow$ &$e_{s}\downarrow$ &$\sigma_{s}\downarrow$ &$e_{f,z}\downarrow$        &MPJPE$\downarrow$ &PA-MPJPE$\downarrow$ &$e_{s}\downarrow$
                &MPJPE$\downarrow$ &PCK$\uparrow$            \\
                \noalign{\hrule height 1pt}
            \hline \hline
            EgoPose~\cite{yuan2019ego}               &130.3  &79.2   &-     &-     &-         &-     &-     &-              &-     &-     \\
            PhysCap~\cite{shimada2020physcap}        &97.4   &65.1   &7.2   &6.9   &-         &107.8   &93.3   &12.2        &104.4        &83.9        \\
            G{\"a}rtner~\etal~\cite{gartner2022trajectory}   &84.0          &56.0          &-          &-         &-         &-     &-     &-              &-     &-     \\ 
            DiffPhy~\cite{gartner2022differentiable}          &81.7         &55.6        &-          &-         &-          &-     &-     &-              &-     &-     \\ 
            SamCon~\cite{liu2015improving}          &78.4    &63.2   &4.0   &4.3   &20.4      &102.4  &95.4  &9.7           &-     &-     \\
            NeuralPhysCap\cite{shimada2021neural}    &76.5   &58.2   &4.5   &6.9    &-        &-     &-      &-    &99.1       &85.5       \\ 
            Neural MoCon~\cite{huang2022neural}      &72.5   &54.6   &3.8   &2.4   &14.4       &93.4  &86.7  &9.2           &-     &-     \\ 
            PoseTriplet~\cite{gong2022posetriplet}   &68.2         &45.1        &-          &-         &-       &- &- &-     &\textbf{79.5}        &\textbf{89.1}        \\ 
            Xie~\etal~\cite{xie2021physics}         &68.1         &-           &4.0        &\textbf{1.3}       &18.9      &-     &-     &-              &-     &-     \\ 
            SimPoE~\cite{yuan2021simpoe}            &56.7         &41.6        &-          &-         &-           &-     &-     &-              &-     &-     \\ 
            D\&D~\cite{li2022d}                  &\textbf{52.5}         &\textbf{35.5}        &-          &-         &-      &-     &-     &-              &-     &-     \\ 
            \textbf{Ours}                       &55.4 &41.3 &\textbf{3.5} &2.1    &\textbf{12.2} &\textbf{79.3} &\textbf{72.8} &\textbf{8.9}  &83.6 &88.1   \\
            
            \noalign{\hrule height 1.5pt}
            \end{tabular}
        }
\caption{Our method can achieve competitive performance in terms of joint accuracy. NeuralPhysCap, PoseTriplet, and D\&D output human motion from the kinematics-based network with implicit physical laws, which may also contain artifacts.}
\label{tab:comparison}
\end{center}
\end{table*}

\subsection{Reverse Diffusion for Motion Capture}\label{sec:motion capture}

Once the networks are trained, we can construct a physics-based human motion capture framework. We first extract image features with a backbone network, and then predict 2D poses for each frame. Then, the latent gaussian distributions can be generated from the extracted image features with the trained VAE encoder. A noisy initial motion is sampled from the distributions for the reverse diffusion process. By applying the physics-based tracking, we can combine the physics information with the image features to guide the denoising. Finally, the physically plausible human motion can be obtained after several iterations.

\section{Experiments}\label{sec:experiment}

\begin{figure*}
    \begin{center}
    \includegraphics[width=1.0\linewidth]{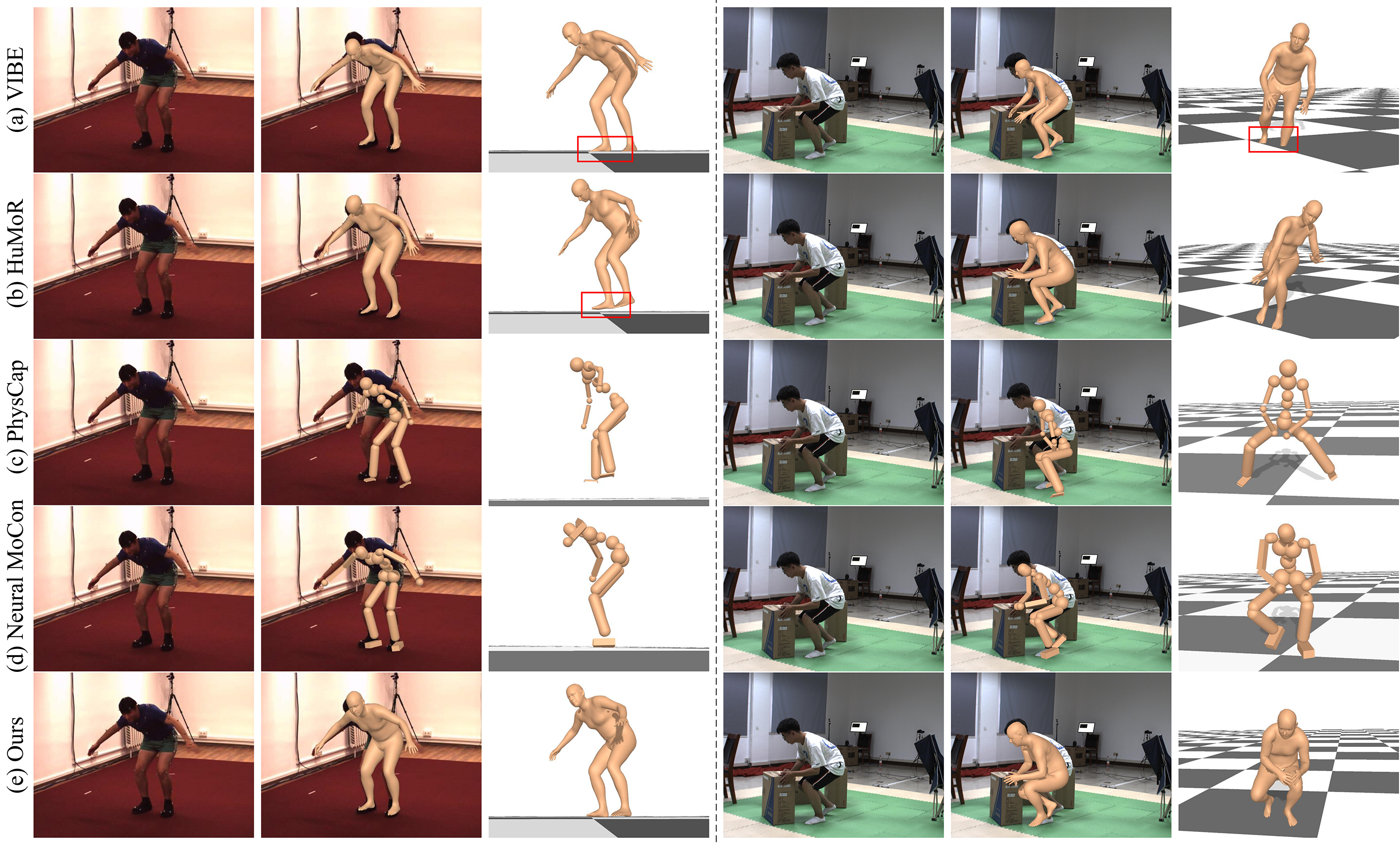}
    \end{center}
    \caption{Qualitative comparison with other methods. The results show that our method can achieve physically plausible and accurate human motion from monocular videos.}
\label{fig:qualitative_comparison}
\end{figure*}

\subsection{Metrics}\label{sec:metrics}
We report the Mean Per Joint Position Error (MPJPE) and the MPJPE after aligning the prediction with ground truth using Procrustes Analysis (PA-MPJPE) to evaluate joint accuracy. We use the 3D extension of the Percentage of Correct keypoints (PCK) at the threshold of 150mm to evaluate the 3D joint position accuracy. We follow PhysCap~\cite{shimada2020physcap} to measure motion jitter error by $e_s$, which is the deviation of joint velocity between predicted output and ground truth. The $e_s$ and its standard deviation $\sigma_s$ are adopted to assess the motion smoothness. To evaluate foot contact, we adopt $e_{f,z}$ proposed in~\cite{xie2021physics}, which is the foot position error on the z-axis.

\subsection{Datasets}\label{sec:datasets}
\noindent\textbf{Human3.6M}~\cite{ionescu2013human3} is an indoor dataset for human motion capture. The videos are captured at 50Hz which includes 7 subjects. Following previous works~\cite{yuan2021simpoe,shimada2020physcap}, we use 2 subjects (S9, S11) for evaluation, and the others are used for training. We convert the dataset to 30Hz to reduce redundancy.

\noindent\textbf{3DOH}~\cite{zhang2020object} is the first dataset to handle the object occluded human body estimation, which contains 3D motions in occluded scenarios. We use the sequence \textit{0013}, \textit{0027}, \textit{0029} to evaluate our method in occlusion cases.

\noindent\textbf{MPI-INF-3DHP}~\cite{mehta2017vnect} is a single-person 3D pose dataset. Following previous works~\cite{gong2022posetriplet,shimada2021neural}, we use its test set to demonstrate the generalization of our method.

\begin{table}
    \begin{center}
        \resizebox{1.0\linewidth}{!}{
            \begin{tabular}{l|c c c c}
            \noalign{\hrule height 1.5pt}
            \begin{tabular}[l]{l}\multirow{1}{*}{Method}\end{tabular}
                &SamCon  &Neural MoCon &UHC &\textbf{Ours}  \\
                \noalign{\hrule height 1pt}
            \hline \hline
            success rate    &76.2\%     &83.4\%     &84.1\%   &89.6\%      \\
            \noalign{\hrule height 1.5pt}
            \end{tabular}
        }
\caption{The success rate on 3DOH dataset. Our method significantly outperforms other physics-based works in terms of success rate.}
\label{tab:success_rate}
\end{center}
\end{table}

\subsection{Comparison with State-of-the-Art Methods}\label{sec:comparison}
We compared our method with state-of-the-art dynamics-based human motion capture approaches on Human3.6M dataset, and their average errors are shown in~\cref{tab:comparison}. All approaches in~\cref{tab:comparison} are dynamics-based methods. EgoPose, SimPoE, and our method rely on RL policy to control the character. When the kinematic motion is inaccurate, EgoPose and SimPoE may fail to track and require re-initialization. In contrast, our method can progressively denoise the artifacts in the reference motion during the reverse diffusion process and promote successful tracking. Thus, our method outperforms these two techniques. In addition, we found that D\&D can achieve the best performance in terms of MPJPE and PA-MPJPE. However, D\&D uses a kinematics-based network to implicitly learn the physical laws, from which the output motion may still contain artifacts. We also use other metrics like motion smoothness and foot contact error to measure the physical plausibility. The results show that our method achieves state-of-the-art on most of the metrics. We further conduct a qualitative comparison with both kinematics-based and physics-based methods in \cref{fig:qualitative_comparison}. VIBE~\cite{kocabas2020vibe} and HuMoR~\cite{rempe2021humor} are recent works that exploit temporal information to obtain kinematic motions. The results show that the reconstructed motion in 3D scenes cannot get accurate contact. Besides, PhysCap is an optimization-based method that can obtain almost physically plausible human motion. However, it also does not consider the scene interactions and cannot utilize 2D observations in physics-based tracking. Thus, the results may hover above the floor and deviate from 2D poses.

We further conduct experiments on 3DOH dataset. Since the dataset contains a lot of object occlusions, it is more difficult to reconstruct an accurate motion. In \cref{fig:qualitative_comparison}, VIBE and HuMoR cannot get satisfactory results due to the depth ambiguity and occlusions. Although Neural MoCon outputs more precise joint positions, it uses a skeletal character and cannot reconstruct accurate body contact. PhysCap, SamCon, and Neural MoCon formulate the physics-based motion capture as a trajectory optimization, and the tracking results strongly depend on the quality of the estimated reference motion. In contrast, our method can adjust both the kinematic and physical motion in the reverse diffusion process. The quantitative results in \cref{tab:comparison} show that our method significantly outperforms these baseline methods in all metrics.

\begin{figure*}
    \begin{center}
    \includegraphics[width=1.0\linewidth]{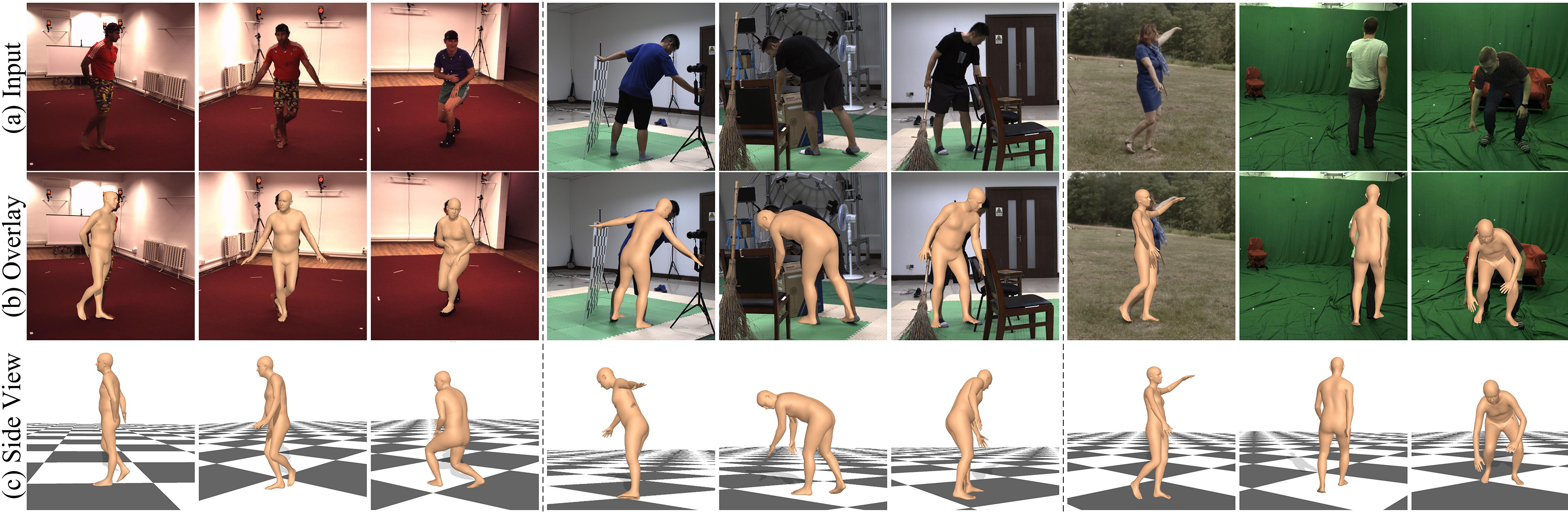}
    \end{center}
    \caption{Qualitative results on Human3.6M, 3DOH, and 3DHP dataset. Our method can produce physically plausible human motion with accurate contact in different scenes.}
\label{fig:result}
\end{figure*}

To evaluate the generalization of our method, we use 3DHP dataset as a benchmark. The results in \cref{fig:result} show that our method can obtain accurate motion with precise contact on different scenes. We also compared our method with PhysCap, NeuralPhysCap, and PoseTriplet on 3DHP. The results in \cref{tab:comparison} show that our method gets more accurate joint positions than PhysCap and NeuralPhysCap, but inferior to PoseTriplet. The reason is that PoseTriplet trains the kinematics-based network assisted by physics but discards the physics module in the pose estimation. Thus, the results from the trained network may also contain artifacts.

\subsection{Ablation Study}\label{sec:ablation}

\begin{table}
    \begin{center}
        \resizebox{1.0\linewidth}{!}{
            \begin{tabular}{l|c c c c}
            \noalign{\hrule height 1.5pt}
            \begin{tabular}[l]{l}\multirow{1}{*}{Method}\end{tabular}
                &step = 1 &step = 5 &step = 10 &step = 50  \\
                \noalign{\hrule height 1pt}
            \hline \hline
            standard w/o tracking    &180.3     &68.3     &43.5   &41.3      \\
            VAE w/o tracking         &55.5   &40.2   &40.1    &39.8      \\
            standard + phys (s=3)    &--     &77.6     &47.1   &43.7          \\
            VAE + phys (s=3)         &--   &41.3   &42.1    &42.8            \\
            \noalign{\hrule height 1.5pt}
            \end{tabular}
        }
\caption{Ablation on the latent gaussian distribution on Human3.6M dataset. The standard gaussian distribution requires more denoising steps to obtain a satisfactory motion, while the latent gaussian distribution can directly sample a plausible motion. The numbers are PA-MPJPE in \textit{mm}.}
\label{tab:vae}
\end{center}
\end{table}

\paragraph{Latent Gaussian Distribution.} The conventional diffusion model requires a lot of denoising steps to reconstruct a satisfactory motion, which is inefficient for the motion capture task. To facilitate the reverse diffusion process, we propose a VAE-based latent gaussian distribution to employ the motion prior knowledge for a good initial value. In \cref{tab:vae}, we compared the proposed latent distribution with the standard gaussian distribution in our motion capture framework. The results show that we can directly sample plausible motions from the encoded distributions. Although the sampled motion may contain a lot of artifacts due to the depth ambiguity, it still has a relatively high joint accuracy. In contrast, we require more denoising steps to reconstruct a motion when we use the standard gaussian as the initial distribution. In the first several steps, the sampled results are pure noises. The framework needs more than 10 steps to reconstruct the motion.

\begin{table}
    \begin{center}
        \resizebox{1.0\linewidth}{!}{
            \begin{tabular}{l|c c c}
            \noalign{\hrule height 1.5pt}
            \begin{tabular}[l]{l}\multirow{1}{*}{Method}\end{tabular}
                &PA-MPJPE$\downarrow$ &$e_s$ $\downarrow$ &success$\uparrow$  \\
                \noalign{\hrule height 1pt}
            \hline \hline
            VAE w/o tracking                    &58.6 &19.8      &--            \\
            VAE w/ tracking                     &66.9       &13.3      &36.5\%           \\
            VAE w/o tracking + denoise (T=1)    &55.5 &20.0   &--            \\
            VAE w/o tracking + denoise (T=3)    &46.2       &14.3   &--            \\
            VAE w/o tracking + denoise (T=5)    &40.2       &16.1   &--            \\
            VAE + phys (s=1) + denoise (T=5)    &43.7   &9.9    &66.7\%            \\
            VAE + phys (s=2) + denoise (T=5)    &41.6   &4.7    &83.4\%            \\
            VAE + phys (s=3) + denoise (T=5)    &41.3   &3.5    &90.3\%            \\
            VAE + phys (s=5) + denoise (T=5)    &41.4   &3.1    &90.7\%            \\
            VAE + phys (s=3) + denoise (T=7)    &41.1   &3.4    &90.9\%            \\
            VAE + phys (s=3) + denoise (T=10)   &42.1   &3.3    &90.0\%           \\
            \noalign{\hrule height 1.5pt}
            \end{tabular}
        }
\caption{Ablation studies. We study the physical guidance with different denoising steps. The tracking denotes applying physics-based tracking, and phys means physical guidance. T denotes the number of timesteps used in the reverse diffusion process, and s is the number of timesteps in which physical guidance is applied. VAE denotes that the reverse diffusion process starts from the latent gaussian distributions.}
\label{tab:physics}
\end{center}
\end{table}

\paragraph{Physical Guidance.} We study the physical guidance in this section. The conventional kinematics-based motion capture predicts the 3D motion from pure image features, which cannot avoid the artifacts due to the depth ambiguity. In \cref{tab:physics}, although the PA-MPJPE of the results sampled from the latent distribution is 58.6, it contains a lot of jitters. We can find that the diffusion model with only kinematics cannot remove the artifacts. Existing physics-based motion capture frameworks use physics as post-processing after the kinematics-based prediction. To demonstrate its weakness with a fair setting, we directly add physics-based tracking on the sampled motion. This strategy can alleviate the artifacts, but it has a low success rate since the noises in the kinematic motion always result in tracking failure. We further use the denoising framework to refine the kinematic motion. The results in \cref{tab:physics} show that the kinematics-based diffusion model can improve the joint accuracy, but does not prevent the artifacts. By incorporating the physical guidance, the success rate can be significantly improved. In addition, the results in PA-MPJPE also demonstrate that the physics also provides the correct direction to enhance the motion capture accuracy. To compare the strategy adopted in PhysDiff, we remove the projection loss gradient condition, and use the tracked motion for the input of the next denoising step in \cref{tab:guide}. We found that the gradients can provide implicit guidance for the diffusion model to optimize the kinematic motion, and thus improve the success rate in the physics-based tracking.

\paragraph{Denoising Step.} We also study the impact of the denoising step. In \cref{tab:physics}, we found that the joint accuracy can be significantly improved with the physical guidance. The performance increases at first with more times of physical guidance and then becomes stable. In addition, the gains are also declining with more than 7 denoising steps.

\begin{table}
    \begin{center}
        \resizebox{1.0\linewidth}{!}{
            \begin{tabular}{l|c c c}
            \noalign{\hrule height 1.5pt}
            \begin{tabular}[l]{l}\multirow{1}{*}{Method}\end{tabular}
                &PA-MPJPE$\downarrow$ &$e_s$ $\downarrow$ &success$\uparrow$  \\
                \noalign{\hrule height 1pt}
            \hline \hline
            phys (s=3) w/o guidance &44.7   &5.9   &73.9\%            \\
            phys (s=3) w/ guidance  &41.3   &3.5   &90.3\%            \\
            \noalign{\hrule height 1.5pt}
            \end{tabular}
        }
\caption{Ablation study on the physical guidance. w/o guidance means that we remove the projection loss gradients in the condition.}
\label{tab:guide}
\end{center}
\end{table}

\section{Conclusion}\label{sec:Conclusion}
In this work, we formulate the physically plausible human motion capture as a reverse diffusion process. Latent gaussian distributions are built based on VAE to utilize the motion prior knowledge to facilitate the reverse diffusion. To employ physics information, we further construct a physics module to combine the physics and image observations to guide the denoising. With the physical guidance, physics and kinematics can promote each other to progressively reconstruct a high-quality human motion. Experimental results on several human motion capture datasets demonstrate that our method can achieve competitive performance with a higher success rate.

\section*{Acknowledgments}
This work was supported in part by the National Natural Science Foundation of China (No. 62076061) and the Natural Science Foundation of Jiangsu Province (No. BK20220127). The authors would like to thank Yuan Yang for the helpful discussion about physics-based imitation.

\section*{Contribution Statement}
Jingyi Ju and Buzhen Huang contribute equally to this work. The corresponding author is Yangang Wang.

\bibliographystyle{named}
\bibliography{ijcai23}

\end{document}